# Real-time Lane detection and Motion Planning in Raspberry Pi and Arduino for an Autonomous Vehicle Prototype

Alfa Rossi
Department of Electrical and
Electronic Engineering
Islamic University of Technology (IUT)
Gazipur, Bangladesh
alfarossi@iut-dhaka.edu

Nadim Ahmed
Department of Electrical and
Electronic Engineering
Islamic University of Technology (IUT)
Gazipur, Bangladesh
nadimahmed@iut-dhaka.edu

Sultanus Salehin
Department of Electrical and
Electronic Engineering
Islamic University of Technology (IUT)
Gazipur, Bangladesh
sultanussalehin@iut-dhaka.edu

Tashfique Hasnine Choudhury
Department of Electrical and
Electronic Engineering
Islamic University of Technology (IUT)
Gazipur, Bangladesh
tashfiquehasnine@iut-dhaka.edu

Golam Sarowar
Department of Electrical and
Electronic Engineering
Islamic University of Technology (IUT)
Gazipur, Bangladesh
asim@iut-dhaka.edu

*Abstract*— **This paper discusses a vehicle prototype that recognizes streets' lanes and plans its motion accordingly without any human input. Pi Camera 1.3 captures real-time video, which is then processed by Raspberry-Pi 3.0 Model B. The image processing algorithms are written in Python 3.7.4 with OpenCV 4.2. Arduino Uno is utilized to control the PID algorithm that controls the motor controller, which in turn controls the wheels. Algorithms that are used to detect the lanes are the Canny edge detection algorithm and Hough transformation. Elementary algebra is used to draw the detected lanes. After detection, the lanes are tracked using the Kalman filter prediction method. Then the midpoint of the two lanes is found, which is the initial steering direction. This initial steering direction is further smoothed by using the Past Accumulation Average Method and Kalman Filter Prediction Method. The prototype was tested in a controlled environment in real-time. Results from comprehensive testing suggest that this prototype can detect road lanes and plan its motion successfully.**

I. INTRODUCTION

Recent years have seen increasing interest in designing and testing of autonomous cars, i.e., vehicles that can navigate without human input, and surveillance vehicles. A fascinating area of research is to build a system that can track road lanes and plan its motion duly. Effective integration of autonomous cars can guarantee less traffic on the streets and help the disabled and venerable travel places without any trouble [2]. Real-time monitoring of road lanes can help us to build autonomous transports though there are other factors to concern before sending out autonomous vehicles. Security against the potential hacking of autonomous cars is nowadays a popular research area. Also, a lot of drivers can lose their job due to the automation of transports. That being said, autonomous vehicle and surveillance is a major research area and these researches can potentially save millions of lives in the future.

In our paper, we designed a prototype of a real autonomous vehicle and took some performance benchmarks. Our paper is organized as follows. Firstly, there is a brief survey of some earlier and contemporary research. It is followed by a description of the hardware configuration of our prototype. Then the methodology is discussed in details, which include image data collection and preprocessing using Gaussian blurring and line detection using Hough transformation, tracking of the lanes by using Kalman filter, calculation of our current position and steering direction, communication between Raspberry Pi and Arduino, planning the motion of the prototype and finally steering the prototype. After that, the performance of our algorithms and the prototype is discussed. Conclusions are presented, and future scope of research is discussed in the final section.

II. PRIOR RESEARCH

A plethora of techniques for lane detection based on visual data has been developed over the years. One of the earliest yet most convenient ways to do it is to apply a Canny edge detector [3] to the image frames and then to apply Hough transform [4] to extract some probable lane markers [1]. Wang, Shen, and Teoh [5] designed a real-time Catmull–Rom spline-based model, which works on setting control points form random shapes. It is introduced along with detection algorithms similar to Hough transformation and maximum likelihood approach using Visual C++ programming software. The most significant part is that the technique functions appropriately for both marked and unmarked lanes due to the system's robustness to noises and visual variations in the images. Borker, Hayes, and Smith [6] used the RANSAC algorithm to eliminate the outliers found from Hough transformation and then took the help of the Kalman filter for tracking the lane markers in case of occasional false-positive detection in some image frames. Horak, and Zalud [7] adopted a real-time vision approach, including Sobel operator, Harris & Stephens operator, and Hough transform algorithm for the edge, corner, and lane detection which is implemented on "Cube," a miniature robot built from scratch. Processes were executed on a Raspberry Pi platform via MATLAB Simulink and tested on four separate resolutions.

Research on lane detection in single board computers has been seen in recent years as well. Kulyukin, and Sudini [8] built an embedded lane detection system on Raspberry Pi based on the detection of one dimensional Haar wavelet spikes. They achieved twenty frames per second processing rate, although the accuracy of the detection of both lanes was below 70%, and a single lane was below 90%. A novel approach to detect lane markers and traffic light images, red color in particular, has been presented by Kalms, Rettkowski, Hamme, and Göhringer [9] using MATLAB Simulink Support Package for Raspberry Pi Hardware to implement algorithms. A camera incorporated with a car model is used for vision purposes. As the proposed method lacks robustness to different lighting conditions, introducing a pattern recognition method and a Neural network was proposed for future researches. Mandlik, and Deshmukh [10] introduced a Lane Departure Warning System (LDWS) and a Lane-Keeping System (LKS), which are significant parts of advanced driver assistance system (ADAS); of which the first one alerts the driver about the departure of the vehicle while the later autonomously keeps the vehicle on the lane using Canny edges detector and Hough transform algorithm. The outcome is efficient lane detection and tracking for straight lanes and curved roads. Lee et al. [11] proposed a Lane Departure Warning System (LDWS) implemented on the IMX6Q board and mounted the system on a real vehicle. A Simple filter was used for doing binarization, and the Kalman filter was used for tracking the lanes, which were found by the Hough transform algorithm. They achieved a 96% detection rate with a processing speed of 15 frames per second.

## III. HARDWARE

A plywood body of $32cm \times 35cm \times 12cm$ houses all the components. Plywood ensures proper isolation of electrical equipment. The plywood was cut using CNC cutter for better measurement accuracy, so it was easy to waterproof the vehicle. Raspberry Pi Model 3 B serves as the primary image processing unit, whereas Pi Cam 1.3 is used for real-time video capture. Arduino Uno Rev3 is used for motion planning and steering unit. 12-volt, 300 RPM, ten kg-cm torque motors are used which are connected to 120 mm diameter wheels. VNH2SP30 full-bridge motor drivers are used to control the motors. Three cell 3300 mAh lithium polymer battery is used and sufficient to power the system. Fig.1 illustrates the hardwire setup.

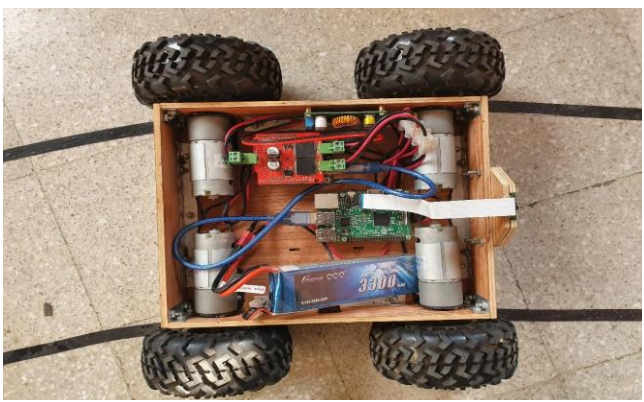

Fig. 1. The Hardware Configuration of the Prototype.

## IV. METHODOLOGY

Raspberry pi is the primary processing unit of our system. Image frames captured by Pi cam goes through different algorithms which help us to find out the lane markers. The python OpenCV library is used to implement the image processing algorithms. Then, the current position of the prototype is determined, and the steering direction is planned. PID algorithm controls the motor speed and consequently controls the direction of motion. Various components of our system are described in details in the following sections.

### A. Data Collection and Preprocessing

The video captured by Pi Camera at the resolution 640 x 480 is resized to half of its width and height; thus, the resolution is converted to 320 x 240. The image frames of the video must be converted to a grayscale image for the upcoming stages of the presented algorithm, namely edge detection and Hough transformation. Gaussian filter is implemented for image blurring, which removes noise. Then canny edge detector is used for edge detection. As it can safely be assumed that the lanes will be situated at the bottom half of the frames, the top half of every frame is masked out so that no unnecessary lines in that region gets detected. The untouched region down below is our region of interest (ROI). The results of preprocessing algorithms can be seen in Fig. 2.

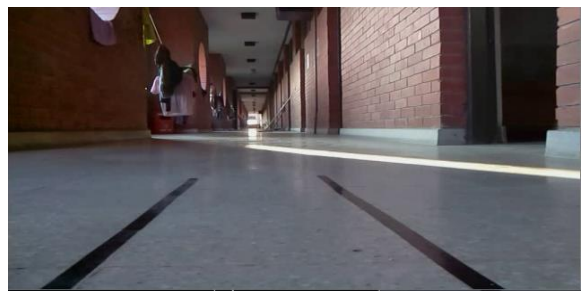

(a) Actual frame.

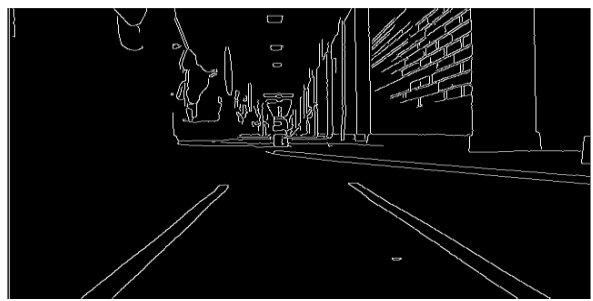

(b) Edge Detected Frame.

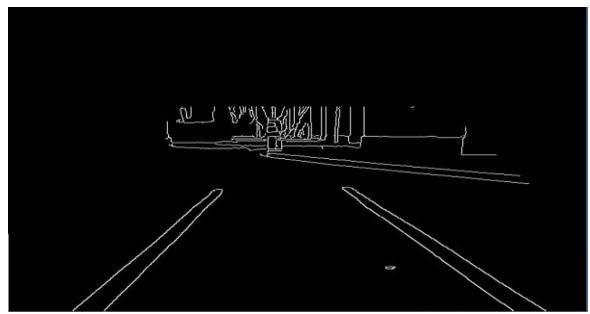

(c) Segmented Frame.

Fig. 2. Preprocessing Done on The Captured Image

## B. Hough Lines Detection

The lines in the preprocessed image frames are found by using Hough Transform on the frames [12,13]. Two parameters are needed to find out the position of the lane marker on the polar axis. One the one hand, the perpendicular distance of the lane markets from the origin (0,0) pixel of our image, which is denoted by ρ, is needed. On the other hand, the angular distance between the lane marker and the vertical coordinate axis is also needed, and it is denoted by ϴ. Hough transformation algorithm gives us these two essential parameters. As the camera is fixed, lines having slopes, ϴ, between -25 to -50 degrees are categorized as left lanes and, lines having slopes between 25 to 55 degrees are categorized as right lanes. By averaging the slope and y-intercept values found from Hough transform, all the lines of the left part are converted to a single left lane and all the lines of the right part is converted to a single right lane. The detected lane is marked in red color and illustrated in Fig. 3.

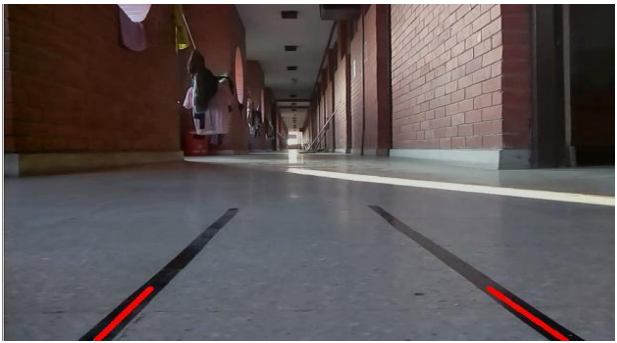

Fig. 3.    Lane Markers Found by Hough Transform.

## C. Line Tracking

It is not uncommon to find image frames in the video taken by Pi camera without any detected lane or with false-positive detection. Though the system recovers very soon, we still have to track the lane markers in a sophisticated way so that the momentary erroneous detections can be mitigated. Kalman filter [14] is an excellent choice to predict the position of the lane markers. The lane trackers orientation parameter, denoted by (ρ, ϴ) is predicted by Kalman filter based on its value on previous frames [6,11]. The tracked lines (red) using Kalman filter is illustrated in Fig. 4.

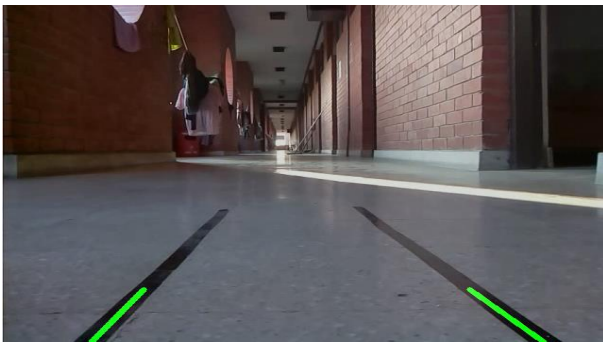

Fig. 4.    Line Markers Being Tracked by Kalman Filter.

## D. Calculating Current Position

The average of the X coordinates of the farthest most end of the two lanes (where the Y-axis coordinate is the ceiling of our ROI) is used to calculate and obtain the vehicle's current position. Since the Pi Cam is mounted precisely at the center of the prototype, the prototype's desired position is at the center of the obtained frame. The setpoint is the 160$^{th}$ pixel of the X-axis's 320 pixels, while the Y coordinate is fixed at the ceiling of ROI. This coordinate should always be the desired position of the vehicle. Initially it is also the prototype's position. When the prototype moves, the current position gets away from the setpoint.

However, there are some shortcomings in finding the current position. Depending on the weather and light condition, there can be some inconsistency in this process. For example, sometimes, the current position can be determined without any complications, and sometimes the current position cannot be readily determined. Two different algorithms were used to overcome this: Past Accumulated Average Method and Kalman Filter Algorithm.
In the Past Accumulated Method, an accumulator matrix of 'n' size is initialized. The past n values of prototype positions are stored sequentially. For our case, we took n = 8, and the matrix elements are averaged to obtain the current position. This algorithm helps us remove the system's failure even if the system misses the lanes for a few frames.
Another algorithm used is the Kalman Filter prediction algorithm, where prototype's current position is predicted based on its previous position. Fig. 5 illustrates the current position based on both methods.

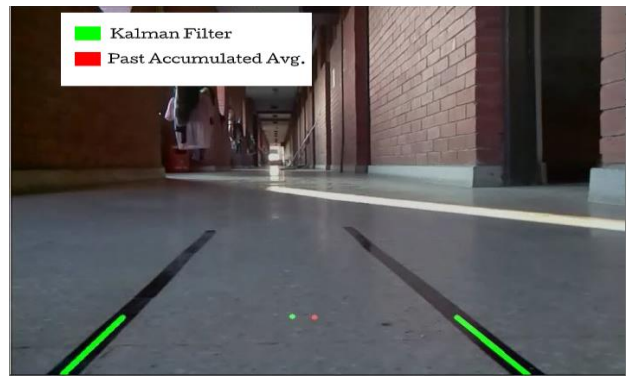

Fig. 5.    Current Position Based on Kalman Filter Prediction Method and Past Accumulated Average Method.

## E. Communication

The current position is sent to the Arduino from the Raspberry Pi using UART communication. The transmitting UART converts parallel data from the Raspberry Pi into serial form and transmits it in serial to the receiving UART, the Arduino. The value of the current position varies from 0 to 320. This three-byte information takes many resources while sending it to the Arduino, which hurts our prototype's real-time performance. So, this range is mapped to a lower range: 1 to 99. Therefore, real-time serial communication is implemented by restricting the data size within two bytes. The current position information is sent only when there is a transition in the value. An indicator ASCII value is sent after transmitting the data of each frame so that the Arduino can differentiate the individual frames' data. The Receiving UART, i.e., Arduino Uno, receives the first byte of serial data and shifts its position by one decimal place and then receives the second byte and keeps on doing this till it gets the predefined ASCII value. As soon as it finds the predefined ASCII value, it breaks the loop and determines the accumulated two-byte value, which corresponds to the vehicle's current position.

*F. Motion Planning*

PID algorithm is implemented on the Arduino Uno. The algorithm calculates the error by subtracting the obtained current position value from time to time from the set point. This error value is used to plan the vehicle's motion using a PID algorithm [15]. The PID algorithm has been tuned by fixing integral constant, derivative constant and proportional constant by trial and error method.

*G. Steering*

Depending on the PID output, the drive function adds the error value of PID to the base value, which is 100 PWM for our case, of left motor and subtracts it from the right or vice-versa depending on the necessity. The highest PWM value is predefined and would not be exceeded. Thus, the vehicle moves smoothly without any sharp turns or extraordinarily high or low PWM and moves autonomously, maintaining the lanes.

## V. EXPERIMENTAL RESULTS

The whole detection algorithm was implemented in Raspberry Pi 3B, and it processed close to six frames every second, which is suitable for moderately speeded vehicles. Initially, we sent 3 bytes of data containing our prototype's current position from Raspberry pi to Arduino via UART, which made our system unable to plan its motion in real-time. However, as we changed our approach, which is described in our methodology, and started to send two-byte positional data to the Arduino, the prototype started working fine in real-time. The prototype has completed ten indoor tracks equipped with bumps, sudden disappearance of lane markers, and barely visible lane markers, which mimics scenarios faced in real roads. The tracks' cumulative length is 2 kilometers, and the prototype completed all of those without going out of the lane. Although there were cases of noisy measurements, our prediction-based tracking algorithms helped our prototype to remain on the track.

As we can see from Fig. 6, the past accumulated average method indeed makes the changes in the direction of motion smoother. The averaging of the previous frames makes the system less sensitive to sudden changes in direction due to false positives in our detection phase. So, the system is less prone to failure now. However, it is not without any drawback as we can see a frame-lag when we use the past accumulating average method. This frame-lag can cause a lack of response when a sudden change in direction is needed.

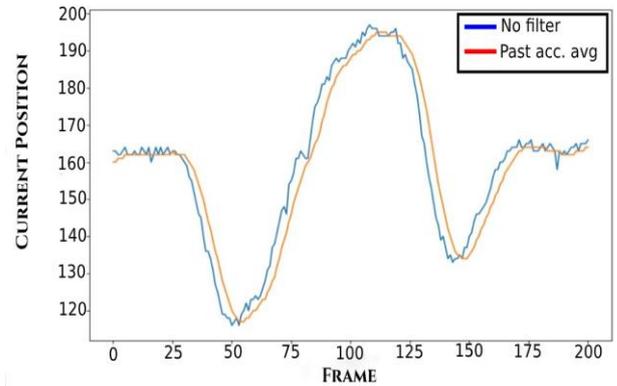

Fig. 6. Comparison Between Current Position Based on Past Accumulated Average Method and only Midpoint of Two lanes.

It is evident from the Fig. 7 that Kalman filter prediction of steering direction does not have any frame-lag, which was present in the past accumulated average method. So, the Kalman filter predicts the movement direction faster. It also makes the sudden changes in direction smoother than the actual result obtained without any filtering, but is not as smooth as the past average accumulation method and is sensitive to sudden changes that can make the system unstable. However, overall, the Kalman filter's prediction of the current position is more convenient to use for us. Our prototype also stops when there is no lane detected for eight frames.

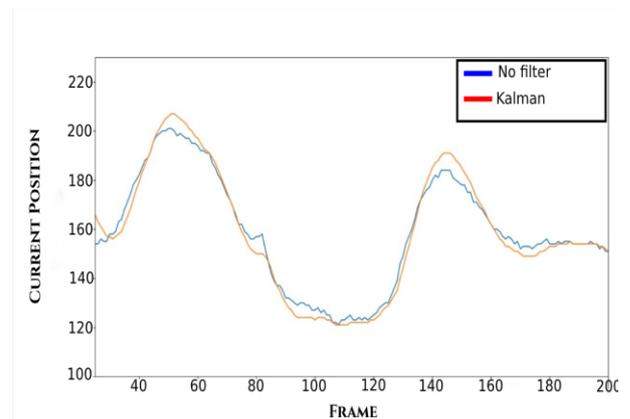

Fig. 7. Comparison Between Current Position Based on Kalman Filter Prediction Method and only Midpoint of Two Lanes.

## VI. CONCLUSION AND FUTURE RESEARCH

This paper implemented a very cost-efficient autonomous vehicle where we used Canny edge detection and Hough Transform to detect road lanes and Kalman filter for tracking our lane markers. The position information was fed from the Raspberry Pi to Arduino via UART communication, which controlled the motors' speed. Several tracks were made on our university campus, mimicking original streets, and our prototype completed all of those. The processing rate was six frames per second.

In future research, the Harris and Stephens corner detection algorithm [16] can be implemented for smoother control on street corners. Using one Raspberry Pi for preprocessing and feeding the preprocessed data to another Raspberry Pi will improve the processing rate. Ackerman Steering [17] can be implemented instead of PWM control.